\documentclass{article}
\usepackage{spconf,amsmath,graphicx,hyperref}
\usepackage{multirow}
\usepackage{multicol}
\usepackage{booktabs}

\title{CS3-Bench: Evaluating and Enhancing Speech-to-Speech LLMs for Mandarin-English Code-Switching}
\name{Heyang Liu$^{1,2}$, Yuhao Wang$^{1,2}$, Ziyang Cheng$^{1}$, Ronghua Wu$^{2}$, Qunshan Gu$^{2}$, Yanfeng Wang$^{1}$, Yu Wang$^{1\dagger}$\thanks{$^\dagger$Corresponding author.}}
\address{$^1$Shanghai Jiao Tong University \quad $^2$ Ant Group}

\begin{document}
%
\maketitle
\begin{abstract}
The advancement of multimodal large language models has accelerated the development of speech-to-speech interaction systems. While natural monolingual interaction has been achieved, we find existing models exhibit deficiencies in language alignment. In our proposed \textbf{C}ode-\textbf{S}witching \textbf{S}peech-to-\textbf{S}peech \textbf{Bench}mark (CS3-Bench), experiments on 7 mainstream models demonstrate a relative performance drop of up to 66\% in knowledge-intensive question answering and varying degrees of misunderstanding in open-ended conversations. Starting from a model with severe performance deterioration, we propose both data constructions and training approaches to improve the language alignment capabilities, specifically employing Chain of Recognition (CoR) to enhance understanding and Keyword Highlighting (KH) to guide generation. Our approach improves the knowledge accuracy from 25.14\% to 46.13\%, with open-ended understanding rate from 64.5\% to 86.5\%, and significantly reduces pronunciation errors in the secondary language. CS3-Bench is available at \url{https://huggingface.co/datasets/VocalNet/CS3-Bench}.



\end{abstract}
\begin{keywords}
Code-switching, speech-to-speech interaction, large language model, evaluation
\end{keywords}

\vspace{-4mm}
\section{Introduction}
\label{sec:intro}
\vspace{-3mm}

Advances in multimodal large language models (LLMs) have enabled speech-capable conversational systems. By expanding the training corpus to include multiple languages, leading open-source models support vocal interactions in both Mandarin and English~\cite{zeng2024glm, xu2025qwen2, ding2025kimi}. Despite the significant progress achieved, we identify a critical challenge concerning language alignment: Can these models maintain high performance when presented with code-switching contexts? Prior research has predominantly focused on speech recognition, neglecting the critical role in speech-to-speech interaction models, which involve both multilingual understanding and coherent, appropriate response generation. 


\begin{figure}[t!]
  \centering
  
  \includegraphics[width=0.35\textwidth]{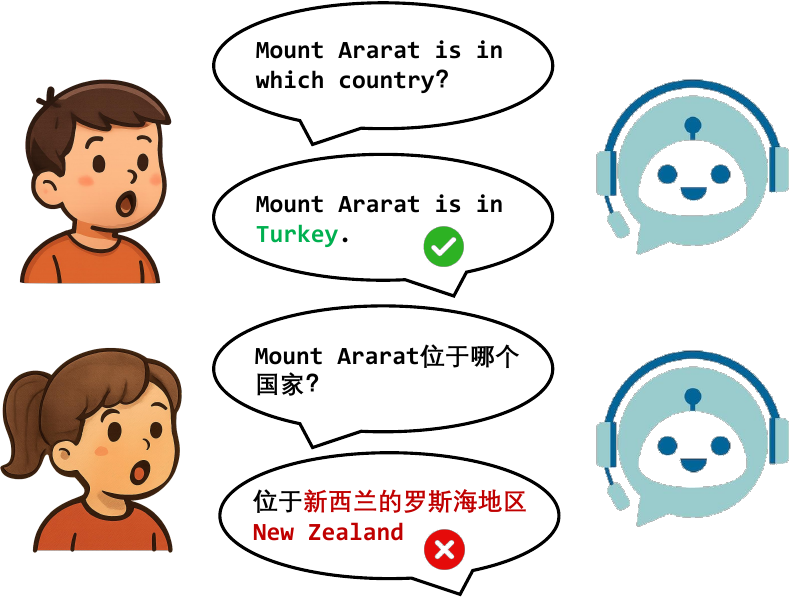}
  \caption{The code-switching deterioration in conversations.}
  \vspace{-7mm}
  \label{fig: cs}
\end{figure}

Code-switching is a prevalent phenomenon in Mandarin conversations, where English words appear as conventional expressions (e.g., ``Cosplay''), named entities of foreign origin (e.g., ``La Quebrada cliff''), and terms associated with academic or technical domains (e.g., ``CRISPR-Cas9''). By developing speech-to-speech interaction models capable of understanding and generating code-switched content, we can enhance the effectiveness and naturalness of voice assistants and further intelligent speech agents. Despite the frequent occurrence, the absence of data resources and evaluation frameworks poses a significant challenge in assessing model performance. While existing benchmarks, such as URO-Bench~\cite{yan2025uro}, include related instances, they often involve only simplistic translations of individual words, failing to capture the natural and contextually grounded patterns of bilingual usage. 



 

To address these difficulties, we propose the \textbf{C}ode-\textbf{S}witching \textbf{S}peech-to-\textbf{S}peech \textbf{Bench}mark (\textbf{CS3-Bench}), to evaluate the performance of speech interactive models with queries predominantly Mandarin and mixed English. CS3-Bench comprises two distinct subsets: a knowledge query set with factual answers, and an open-ended set to simulate naturalistic conversations. We establish a comprehensive evaluation framework assessing model performance across semantic quality, pronunciation accuracy, and language selection. Our findings reveal that all evaluated models exhibit a significant decline when transitioning from monolanguage to code-switching contexts, underscoring the challenges in effectively handling mixed-language inputs. Based on a model with severe deterioration, we propose a multi-faceted enhancement strategy, including a data construction pipeline to support code-switching fine-tuning, a chain-of-recognition approach to decompose the understanding process, and keyword highlighting to emphasize the foreign-language terms in generation. Our contributions are summarized as follows:

\begin{table*}[htbp]
\centering
\resizebox{1\textwidth}{!}{
\begin{tabular}{ccccccccccccccccccccc}
\toprule

 \multirow{3}{*}{\textbf{Model}}  & \multicolumn{8}{c}{\textbf{Knowledge}} & \multicolumn{5}{c}{\textbf{Open-ended}} \\  \cmidrule(lr){2-9} \cmidrule(lr){10-14} 
 
 & \multicolumn{4}{c}{\textbf{Semantic Accuracy (\%)}} & \multicolumn{3}{c}{\textbf{Pronunciation}} & \textbf{Language} & \multicolumn{2}{c}{\textbf{Semantic}} &\multicolumn{3}{c}{\textbf{Pronunciation}} \\ \cmidrule(lr){2-5} \cmidrule(lr){6-8} \cmidrule(lr){9-9} \cmidrule(lr){10-11} \cmidrule(lr){12-14}  
 
 & Acoustic & Semantic & Integrated & Avg & Num & PSR (\%) & WER (\%) & LSA (\%) & Understanding (\%) & Generation & Num & PSR (\%) & WER (\%)  \\ \midrule

 VocalNet-ML~\cite{wang2025vocalnet} &  25.14 (75.41) & 29.66 (80.51) & 22.95 (75.41) & 26.24 (77.07) & 103 & 62.14 & 83.50 & 99.45 & 64.5 & 3.42 & 269 & 51.30 & 53.16 \\

 GLM-4-Voice~\cite{zeng2024glm}  & 30.60 (59.56) & 22.88 (62.71) & 31.15 (65.57) & 28.18 (61.60) & 1422 & 75.25 & 26.58 & 96.69 & 73.5 & 3.80 & 639 & 65.88 & 48.83 \\

 VITA-Audio-Plus-Vanilla~\cite{long2025vita} & 37.70 (48.09) & 33.90 (46.61) & 40.98 (59.02) & 37.02 (49.45) & 1191 & 58.86 & 44.33 & 99.45 & 74.5 & 3.89 & 1068 & 77.81 & 25.75 \\

 Baichuan-Omni-1.5~\cite{li2025baichuan} & 38.25 (67.21) & 38.14 (66.95) & 36.07 (70.49) & 37.85 (67.68) & 1526 & 65.01 & 37.02 & 99.45 & 82.5  & 3.72 & 2124 & 82.77 & 20.72 \\

 Qwen2.5-Omni~\cite{xu2025qwen2} & 37.16 (77.60) & 44.92 (76.27) & 50.82 (75.41) & 41.99 (76.80) & 685 & 79.42 & \textbf{23.65} & 95.86 & 83.0  & 3.83 & 531 & \textbf{86.63} & \textbf{17.70} \\

 MiniCPM-o 2.6~\cite{MiniCPM-o-2.6} & \textbf{52.46} (\textbf{81.42}) & 50.00 (77.12) & 50.82 (80.33) & 51.38 (79.83) & 1291 & \textbf{81.95} & 25.33 & 95.86 & \textbf{87.5} & \textbf{4.08} & 995 & 85.63 & 46.43 \\

 Kimi-Audio~\cite{ding2025kimi} & 50.27 (79.23) & \textbf{52.54} (\textbf{81.36}) & \textbf{60.66} (\textbf{81.97}) & \textbf{52.76} (\textbf{80.39}) & 663 & 53.85 & 49.47 & \textbf{100} & 81.0  & 3.83 & 532 & 65.86 & 38.16 \\
\bottomrule
 \vspace{-5mm}
\end{tabular}
}
\caption{Evaluation performance on CS3-Bench. The numbers in brackets are the accuracy on the English version.}
\vspace{-5mm}
\label{tab: cs3bench_performance}
\end{table*}

\begin{itemize}
    \item We propose CS3-Bench, the first code-switching benchmark specifically designed for speech interaction scenarios, and develop appropriate evaluation metrics.
    \item We evaluate 7 leading interactive speech LLMs and observe significant performance degradation.
    \item We propose training strategies from both understanding and generation perspectives, and present an efficient framework for training data generation and fine-tuning.
\end{itemize}




\begin{figure*}[h]
  \centering
 \includegraphics[width=0.88\textwidth]{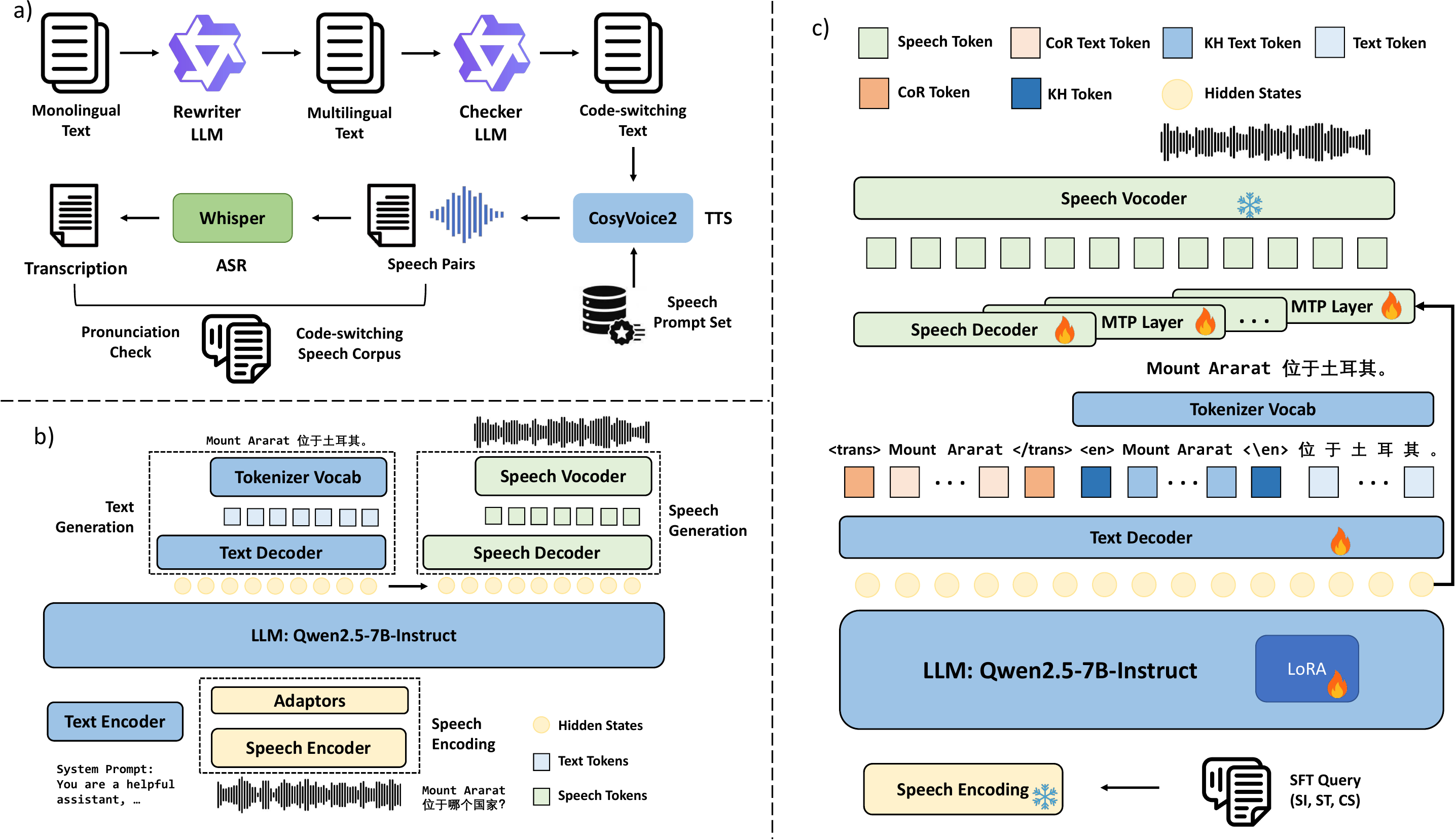}
 \vspace{-3mm}
  \caption{The methods for code-switching in interactive speech LLMs. a) The creation pipeline for the code-switching corpus. b) The model architecture for VocalNet baseline. c) The training strategy with the chain of recognition and keyword highlighting.}
  \vspace{-5mm}
  \label{fig: methods}
\end{figure*}

\vspace{-5mm}
\section{CS3-Bench}
\label{sec:benchmark}
\vspace{-2mm}
\subsection{Creation Pipeline}


CS3-Bench is developed as a Mandarin-based evaluation corpus for speech interaction models, with each instance containing specific English lexical items. The instances in the knowledge sets are selected from VocalBench~\cite{liu2025vocalbench}. We initially translate into Chinese using Qwen2.5-Max~\cite{Yang2024Qwen25TR}, and manually retain only those with challenges to formulate purely in Mandarin. The instances are classified into three categories: acoustic, semantic, and integrated. Acoustic instances primarily involve proper names with salient phonetic similarities. Semantic instances are based on conceptual or meaning-based alignments. The integrated category comprises instances that exhibit both acoustic and semantic characteristics. For the open-ended component, we employ GPT-4o~\cite{hurst2024gpt} to generate simulated queries related to technology, health, and lifestyle, and manually collect instances reflective of daily conversations. These instances authentically capture the code-switching phenomena that commonly occur and ensure a high degree of linguistic and contextual diversity. The speech queries are synthesized using CosyVoice2~\cite{du2024cosyvoice2}, with a randomly selected speech prompt drawn from a clean subset of seedtts-eval-eval~\cite{anastassiou2024seed}. Speech recognition is performed using Whisper-large-v3~\cite{radford2023robust}, and any discrepancies are manually verified and corrected through re-synthesis. Additionally, CS3-Bench includes an English version of the knowledge set, enabling the quantification of model performance in monolingual scenarios, thereby providing comparisons outside of code-switching contexts. The resulting benchmark comprises 362 instances in the knowledge set and 200 in open-ended.

 

\vspace{-3mm}
\subsection{Evaluation Metrics}
\vspace{-2mm}

We quantify model performance across semantic quality, pronunciation clarity, and language selection. For the knowledge set, we calculate the accuracy based on keyword spotting and LLM-as-a-Judge if it fails. With respect to pronunciation clarity, we introduce the pronunciation success rate (PSR), defined as the proportion of English words in responses that are recognized by a speech recognition system. We also report the word error rate (WER) for the English segments. Regarding language selection, the language selection accuracy (LSA) is the proportion of responses in which Mandarin is the dominant language, reflecting the ability to maintain linguistic consistency. For the open-ended set, we decompose the semantic quality into understanding and generation with a 5-point scale using LLM-as-a-Judge. The understanding phase is assessed based on the relevance of the model's response to the input query, and we report the proportion of scores above 3. The generation quality is based on the overall assessment of completeness, wording, details, and naturalness.

\vspace{-5mm}
\subsection{Code-Switching Deterioration}
\vspace{-2mm}


We have evaluated 7 mainstream speech interaction models, and the performance on CS3-Bench is shown in Table~\ref{tab: cs3bench_performance}. On the knowledge set, all evaluated models demonstrated strong performance on monolingual queries. However, a significant decline occurs when the query is shifted to a code-switching setting. Six of the evaluated models exhibited an absolute performance drop of 30\% or more, nearly half of the originally correct instances. The remaining model, VITA-Audio-Plus-Vanilla, also experiences a 12\% absolute decrease in accuracy, corresponding to a relative reduction of 25.1\%. Regarding word pronunciation, with the exception of VocalNet-ML, all other models tend to generate responses with English items; however, pronunciation errors remain prevalent. Although Whisper has inherent limitations in recognizing code-switched content, the results indicate that VocalNet, VITA-Audio, and Kimi-Audio exhibit notable deficiencies in multilingual speech generation. For language selection, all models identify the dominant language in the vast majority of instances, achieving accuracy above 95\%. This metric is therefore omitted in subsequent reporting.


In open-ended sessions, around half models fail to understand over 20\% of user queries and generate improper responses. In the assessment of semantic quality, it shows a clear positive correlation with the degree of query understanding, with MiniCPM-o 2.5 achieving the best performance over these two indicators. For word pronunciation, Qwen2.5-Omni is able to correctly generate 86.63\% of the English words in its pronunciation, with a WER of 17.70\%.




\vspace{-3mm}
\section{Methods}
\label{sec:methods}
\vspace{-3mm}
As shown in Fig~\ref{fig: methods}, to enhance model performance in code-switching scenarios, we implement improvements in both data construction and model training strategies. First, we incorporated multilingual data encompassing speech translation and multilingual interleaving through an efficient speech generation pipeline. This enriched the model’s exposure to mixed-language patterns. Second, we designed training strategies to encourage the model to attend to information conveyed in the accompanying language. 

\vspace{-3mm}


\subsection{Language Alignment Data Construction}

The available fine-tuning speech corpus is predominantly composed of monolingual content. We begin by constructing appropriate code-switching instances. Chinese conversations are sourced from moss-003-sft-data~\cite{Sun2024MOSS} and filtered to retain natural conversational interactions. A Rewriter LLM is employed to convert specific words into English, and a Checker LLM to assess whether the rewritten instances are plausible in daily conversational contexts. Then we employ CosyVoice2 to simulate clear and diverse voices, and the generated speech is transcribed using Whisper-large-v3, and only samples with fully correct English content are retained. Following this procedure, 20\% of the text examples are successfully synthesized and preserved, resulting in 19.4k instances. To enhance multilingual alignment, we also process the speech translation dataset CoVoST~\cite{wang2020covost}. Each entry includes a translation instruction as a prefix, with most in the source language and a few in the target language. This part comprises 12k Chinese-to-English and English-to-Chinese translations, respectively.

\vspace{-3mm}
\subsection{Chain of Recognition}

Recognition errors in English content are a primary cause of incorrect responses, particularly in the knowledge set, where the English content predominantly consists of named entities that appear infrequently in both the base LLM training and speech supervised fine-tuning. As shown in Fig~\ref{fig: methods}, to mitigate this issue, we introduce a pre-task to instruct the model to recognize English content in the user query, named chain of recognition (CoR). The recognized items are chained before the model output, and instruct the model to better understand user queries and thus benefit the response generation. A special token is used to separate the CoR output from the model response tokens, and the recognition content is discarded during inference. This approach is similar to the chain-of-thought method, and we strictly limit the recognition length to a maximum of four individual words to avoid impairing the instruction-following capability. 

\vspace{-3mm}
\subsection{Keyword Highlighting}

\begin{table*}[t!]
\centering
\resizebox{1\textwidth}{!}{
\begin{tabular}{lccccccccccccccccccc}
\toprule

 \multirow{3}{*}{\textbf{Model}}  & \multicolumn{7}{c}{\textbf{Knowledge}} & \multicolumn{5}{c}{\textbf{Open-ended}} \\  \cmidrule(lr){2-8} \cmidrule(lr){9-13} 
 
  & \multicolumn{4}{c}{\textbf{Semantic Accuracy (\%)}} & \multicolumn{3}{c}{\textbf{Pronunciation}} & \multicolumn{2}{c}{\textbf{Semantic}} &\multicolumn{3}{c}{\textbf{Pronunciation}} \\ \cmidrule(lr){2-5} \cmidrule(lr){6-8}  \cmidrule(lr){9-10} \cmidrule(lr){11-13}  
 
 & Acoustic & Semantic & Integrated & Avg & Num & PSR (\%) & WER (\%) & Understanding (\%) & Generation & Num & PSR (\%) & WER (\%)  \\ \midrule

 VocalNet-ML &  25.14 & 29.66 & 22.95 & 26.24 & 103 & 62.14 & 83.50  & 64.5 & 3.42 & 269 & 51.30 & 53.16 &  \\
 + ST data & 26.23 & 33.90 & 27.87 & 29.01 & 285 & 45.26 & 61.40 &  68.5 & 3.57 & 225 & 51.11 & 52.89 \\
 + CS data & 34.97 & 39.83 & 42.62 & 37.85 & 1088 & 59.56 & 43.38 &  \textbf{89.5} & \textbf{4.25} & 1102 & 81.67 & 20.51 \\

  + ST, CS data & 39.34 & 44.07 & 45.90 & 41.99 & 1223 & 61.31 & 43.96 & 86.0 & 4.14 & 1144 & 79.81 & 22.73   \\ \hline

 + CoR & 40.98 & 46.61 & 44.26 & 43.37 & 1420 & \textbf{66.76} & 40.42 & 86.5  & 4.23 & 1087 & \textbf{85.74} & \textbf{15.18} \\
 + KH & \textbf{42.62} & \textbf{50.00} & \textbf{49.18} & \textbf{46.13} & 1549 & 63.01 & \textbf{40.09} & 84.0 & 4.20 & 1119 & 83.82 & 17.16\\
 
\bottomrule
\end{tabular}
}
\caption{Evaluation performance on CS3-Bench with language alignment finetuning and specialized training strategies.}
\label{tab: cs3bench_performance}
\vspace{-5mm}
\end{table*}

When encountering code-switching queries, the model responses often include content in the secondary language to express semantics that are difficult to convey in the primary language. As shown in Fig~\ref{fig: methods}, to emphasize this language transition, we introduce indicative tokens to signal the shift and highlight keywords in the secondary language. In speech generation, keyword highlighting guides the speech decoder to produce speech tokens related to both languages. More specifically, during the training phase, the backbone of the large model is fine-tuned using LoRA and thus instructed to produce text tokens related to the English content, typically identical or semantically related to segments in the user query. The speech decoder is simultaneously trained to generate clear and well-pronounced speech responses.


\vspace{-4mm}

\section{Experiment}
\label{sec:exp}

\vspace{-2mm}
\subsection{Model Architecture and Training Configurations}

As shown in Fig.~\ref{fig: methods}, we incorporate VocalNet-ML, a bilingual model based on Qwen2.5-7B-Instruct, which includes extended speech encoding and speech generation components. Specifically, VocalNet-ML employs multi-token prediction (MTP) to accelerate decoding, as illustrated in Fig~\ref{fig: methods} c). Our training data consists of three parts: 19.4k code-switching (CS) instances, 24k speech translation (ST) samples, and a corresponding amount of speech instruction (SI) data equal to language alignment instances, sourced from VoiceAssistant-400K and Belle~\cite{chen2024slam}. During training, only the speech encoding components and the speech vocoder are frozen. The model training is conducted on 4 NVIDIA A100 GPUs.



\vspace{-3mm}
\subsection{Code-Switching Interation Performance}

After incorporating language alignment data, the performance on CS3-Bench is significantly improved. The inclusion of code-switching data enables the generation of interleaved responses in both languages, while substantially enhancing overall performance. The inclusion of both types enables the model to achieve an accuracy improvement from 26.24\% to 41.99\% on the knowledge set. From this model, we introduce the CoR and KH training methods. Both methods achieve further accuracy improvements and pronunciation promotion, with the latter reaching 46.13\% accuracy with 40.09\% WER. More specifically, greater improvements for KH are observed in semantic and integrated questions, consistent with the discussion in the Methods section. Although the VocalNet-based model still lags behind Qwen2.5-Omni and Kimi-Audio, we attribute this gap largely to disparities in training volume: the former is trained on thousands of hours, while the others typically use millions. This difference is particularly evident in the coverage of rare words in conversations, especially named entities, which frequently appear as answers in knowledge tests.

In the open-ended evaluation, the model with CoR achieves a WER of 15.18\% and a semantic score of 4.23, surpassing all other models, and an understanding rate of 86.5\%, slightly behind MiniCPM-o 2.6. In this evaluation set, the English items are typically common expressions, which confer a performance advantage when training data is limited.

Applying both methods simultaneously causes degradation, probably due to an inconsistency - the recognized and generated English items are consistent in some examples but differ in others, which disrupts the training process.

\vspace{-3mm}

\subsection{Monolingual Interaction Performance}
\begin{table}[h!]
\vspace{-0.2cm}
\centering
\resizebox{0.5\textwidth}{!}{
\begin{tabular}{lccc}
\toprule
 \multirow{2}{*}{\textbf{Model}}  & \multicolumn{3}{c}{\textbf{VocalBench}} \\   \cmidrule(lr){2-4}

 & Knowledge (s2t, \%)  & Knowledge (s2s, \%) & Single-Round  \\ \hline

 VocalNet-ML & 61.55 & 58.05 & 3.43 \\
 + ST, CS data & 70.65 & 67.00 & 3.71 \\ 
 \hline 
 + CoR & \textbf{71.45} & \textbf{67.65} & \textbf{3.78} \\
 + KH  & 71.15 & 66.55 & \textbf{3.78} \\ \midrule 
  \multirow{2}{*}{\textbf{Model}}  & \multicolumn{3}{c}{\textbf{VocalBench-zh}} \\   \cmidrule(lr){2-4}

 & Chinsese (\%)  & Foreign (\%) & General (\%) \\ \hline

 VocalNet-ML & 35.3 & 41.1 & 51.1 \\
 + ST, CS data & \textbf{39.7} & 44.5 & \textbf{53.6} \\ 
 \hline 
 + CoR & 39.4 & \textbf{44.9} & 52.7 \\
 + KH  & 39.3 & 44.6 & 52.8 \\
\bottomrule
\end{tabular}
}
\caption{Monolingual performance with proposed methods.}
\vspace{-3mm}
\label{tab: vocalbench_performance}
\end{table}

\noindent To evaluate the model's performance in monolingual contexts, we conduct tests on the knowledge and single-round sets of VocalBench~\cite{liu2025vocalbench}, and the VocalBench-zh knowledge test comprising three categories: Chinese, Foreign, and General. As shown in Table~\ref{tab: vocalbench_performance}, models trained with CS and ST data achieve improvements across all test sets, and our training scheme consistently improves English performance.

\vspace{-3mm}

\section{Conclusion}
\vspace{-3mm}
In this work, we address the language alignment deficiencies in code-switching contexts for speech-to-speech interaction models. We introduce CS3-Bench, an evaluation framework primarily in Mandarin with English content, and observe significant performance degradation in knowledge queries and varying degrees of misunderstanding in open-ended settings within 7 mainstream models. We propose to alleviate this issue through data construction and training strategies, introducing the chain of recognition (CoR) and keywords highlighting (KH). The improved model achieves substantially better performance on the knowledge set and attains the best results across multiple metrics in open-ended evaluation.

\newpage
\bibliographystyle{IEEEbib}
\bibliography{main}

\begin{thebibliography}{10}

\bibitem{zeng2024glm}
Aohan Zeng, Zhengxiao Du, Mingdao Liu, Kedong Wang, Shengmin Jiang, Lei Zhao, Yuxiao Dong, and Jie Tang,
\newblock ``Glm-4-voice: Towards intelligent and human-like end-to-end spoken chatbot,''
\newblock {\em arXiv preprint arXiv:2412.02612}, 2024.

\bibitem{xu2025qwen2}
Jin Xu, Zhifang Guo, Jinzheng He, Hangrui Hu, Ting He, Shuai Bai, Keqin Chen, Jialin Wang, Yang Fan, Kai Dang, et~al.,
\newblock ``Qwen2.5-omni technical report,''
\newblock {\em arXiv preprint arXiv:2503.20215}, 2025.

\bibitem{ding2025kimi}
Ding Ding, Zeqian Ju, Yichong Leng, Songxiang Liu, Tong Liu, Zeyu Shang, Kai Shen, Wei Song, Xu~Tan, Heyi Tang, et~al.,
\newblock ``Kimi-audio technical report,''
\newblock {\em arXiv preprint arXiv:2504.18425}, 2025.

\bibitem{yan2025uro}
Ruiqi Yan, Xiquan Li, Wenxi Chen, Zhikang Niu, Chen Yang, Ziyang Ma, Kai Yu, and Xie Chen,
\newblock ``Uro-bench: A comprehensive benchmark for end-to-end spoken dialogue models,''
\newblock {\em arXiv preprint arXiv:2502.17810}, 2025.

\bibitem{wang2025vocalnet}
Yuhao Wang, Heyang Liu, Ziyang Cheng, Ronghua Wu, Qunshan Gu, Yanfeng Wang, and Yu~Wang,
\newblock ``Vocalnet: Speech llm with multi-token prediction for faster and high-quality generation,''
\newblock {\em arXiv preprint arXiv:2504.04060}, 2025.

\bibitem{long2025vita}
Zuwei Long, Yunhang Shen, Chaoyou Fu, Heting Gao, Lijiang Li, Peixian Chen, Mengdan Zhang, Hang Shao, Jian Li, Jinlong Peng, et~al.,
\newblock ``Vita-audio: Fast interleaved cross-modal token generation for efficient large speech-language model,''
\newblock {\em arXiv preprint arXiv:2505.03739}, 2025.

\bibitem{li2025baichuan}
Yadong Li, Jun Liu, Tao Zhang, Song Chen, Tianpeng Li, Zehuan Li, Lijun Liu, Lingfeng Ming, Guosheng Dong, Da~Pan, et~al.,
\newblock ``Baichuan-omni-1.5 technical report,''
\newblock {\em arXiv preprint arXiv:2501.15368}, 2025.

\bibitem{MiniCPM-o-2.6}
OpenBMB,
\newblock ``Minicpm-o 2.6: A gpt-4o level mllm for vision, speech, and multimodal live streaming on your phone,'' \url{https://openbmb.notion.site/185ede1b7a558042b5d5e45e6b237da9}, 2025,
\newblock Accessed: 2025-03-28.

\bibitem{liu2025vocalbench}
Heyang Liu, Yuhao Wang, Ziyang Cheng, Ronghua Wu, Qunshan Gu, Yanfeng Wang, and Yu~Wang,
\newblock ``Vocalbench: Benchmarking the vocal conversational abilities for speech interaction models,''
\newblock {\em arXiv preprint arXiv:2505.15727}, 2025.

\bibitem{Yang2024Qwen25TR}
Qwen~An Yang, Baosong Yang, Beichen Zhang, Binyuan Hui, Bo~Zheng, Bowen Yu, Chengyuan Li, Dayiheng Liu, Fei Huang, Guanting Dong, Haoran Wei, Huan Lin, Jian Yang, Jianhong Tu, Jianwei Zhang, Jianxin Yang, Jiaxin Yang, Jingren Zhou, Junyang Lin, Kai Dang, Keming Lu, Keqin Bao, Kexin Yang, Le~Yu, Mei Li, Mingfeng Xue, Pei Zhang, Qin Zhu, Rui Men, Runji Lin, Tianhao Li, Tingyu Xia, Xingzhang Ren, Xuancheng Ren, Yang Fan, Yang Su, Yi-Chao Zhang, Yunyang Wan, Yuqi Liu, Zeyu Cui, Zhenru Zhang, Zihan Qiu, Shanghaoran Quan, and Zekun Wang,
\newblock ``Qwen2.5 technical report,''
\newblock {\em arXiv preprint arXiv:2412.15115}, 2024.

\bibitem{hurst2024gpt}
Aaron Hurst, Adam Lerer, Adam~P Goucher, Adam Perelman, Aditya Ramesh, Aidan Clark, AJ~Ostrow, Akila Welihinda, Alan Hayes, Alec Radford, et~al.,
\newblock ``Gpt-4o system card,''
\newblock {\em arXiv preprint arXiv:2410.21276}, 2024.

\bibitem{du2024cosyvoice2}
Zhihao Du, Yuxuan Wang, Qian Chen, Xian Shi, Xiang Lv, Tianyu Zhao, Zhifu Gao, Yexin Yang, Changfeng Gao, Hui Wang, et~al.,
\newblock ``Cosyvoice 2: Scalable streaming speech synthesis with large language models,''
\newblock {\em arXiv preprint arXiv:2412.10117}, 2024.

\bibitem{anastassiou2024seed}
Philip Anastassiou, Jiawei Chen, Jitong Chen, Yuanzhe Chen, Zhuo Chen, Ziyi Chen, Jian Cong, Lelai Deng, Chuang Ding, Lu~Gao, et~al.,
\newblock ``Seed-tts: A family of high-quality versatile speech generation models,''
\newblock {\em arXiv preprint arXiv:2406.02430}, 2024.

\bibitem{radford2023robust}
Alec Radford, Jong~Wook Kim, Tao Xu, Greg Brockman, Christine McLeavey, and Ilya Sutskever,
\newblock ``Robust speech recognition via large-scale weak supervision,''
\newblock in {\em International conference on machine learning}. PMLR, 2023, pp. 28492--28518.

\bibitem{Sun2024MOSS}
Tianxiang Sun, Xiaotian Zhang, Zhengfu He, Peng Li, Qinyuan Cheng, Xiangyang Liu, Hang Yan, Yunfan Shao, Qiong Tang, Shiduo Zhang, Xingjian Zhao, Ke~Chen, Yining Zheng, Zhejian Zhou, Ruixiao Li, Jun Zhan, Yunhua Zhou, Linyang Li, Xiaogui Yang, Lingling Wu, Zhangyue Yin, Xuanjing Huang, Yu-Gang Jiang, and Xipeng Qiu,
\newblock ``Moss: An open conversational large language model,''
\newblock {\em Machine Intelligence Research}, 2024.

\bibitem{wang2020covost}
Changhan Wang, Juan Pino, Anne Wu, and Jiatao Gu,
\newblock ``Covost: A diverse multilingual speech-to-text translation corpus,''
\newblock in {\em Proceedings of the Twelfth Language Resources and Evaluation Conference}, 2020, pp. 4197--4203.

\bibitem{chen2024slam}
Wenxi Chen, Ziyang Ma, Ruiqi Yan, Yuzhe Liang, Xiquan Li, Ruiyang Xu, Zhikang Niu, Yanqiao Zhu, Yifan Yang, Zhanxun Liu, et~al.,
\newblock ``Slam-omni: Timbre-controllable voice interaction system with single-stage training,''
\newblock {\em arXiv preprint arXiv:2412.15649}, 2024.

\end{thebibliography}

\end{document}